%% file: ijcai24.tex
\documentclass{article}
\usepackage{ijcai24}

\usepackage{times}
\usepackage{soul}
\usepackage{url}
\usepackage[hidelinks]{hyperref}
\usepackage[utf8]{inputenc}
\usepackage[small]{caption}
\usepackage{graphicx}
\usepackage{amsmath}
\usepackage{amsthm}
\usepackage{booktabs}
\usepackage{algorithm}
\usepackage{algorithmic}
\usepackage[switch]{lineno}
\usepackage{lipsum}

\usepackage{accents}
\usepackage[american]{babel}
\usepackage{overpic}
\usepackage{multirow}

\usepackage[utf8]{inputenc} 
\usepackage[T1]{fontenc}    
\usepackage{hyperref}       
\usepackage{url}            
\usepackage{booktabs}       
\usepackage{amsfonts}       
\usepackage{nicefrac}       
\usepackage{microtype}      
\usepackage{xcolor}         
\usepackage{graphicx}
\usepackage{subfigure}
\usepackage{amsmath,amssymb} 
\usepackage{caption} 

\usepackage{wrapfig}
\usepackage[american]{babel}

\usepackage{graphicx} 
\usepackage{algorithm}
\usepackage{algorithmic}
\usepackage{times}
\usepackage{soul}
\usepackage{url}
\usepackage[utf8]{inputenc}
\usepackage{amsmath}
\usepackage{amsthm}
\usepackage{booktabs}
\usepackage{algorithm}
\usepackage{algorithmic}
\usepackage{subfigure}
\usepackage{amsmath}
\usepackage{amssymb}
\usepackage{booktabs}
\usepackage{multirow}
\usepackage{paralist,algorithmic,algorithm}
\usepackage[american]{babel}
\usepackage{microtype}
\usepackage{lipsum}
\usepackage{bm}
\usepackage{overpic}
\usepackage[switch]{lineno}  %
\newcommand{\x}{{\bf x}}
\newcommand{\w}{{\bm w}}

\newcommand{\cc}{{\bm c}}

\newcommand{\X}{\mathcal{X}}
\newcommand{\D}{\mathcal{D}}
\newcommand{\Y}{\mathcal{Y}}

\newcommand{\R}{\mathbb{R}}

\newcommand{\eg}{\emph{e.g.}}
\newcommand{\ie}{\emph{i.e.}}

\usepackage{amsmath}
\usepackage{xcolor,colortbl}

\definecolor{DarkGreen}{RGB}{1,50,32}
\definecolor{dr}{RGB}{47,85,151}
\definecolor{dreg}{RGB}{112,48,160}
\definecolor{dn}{RGB}{255, 117, 143}
\definecolor{pr}{RGB}{228,197,111}
\definecolor{kd}{RGB}{251,133,0}
\definecolor{mr}{RGB}{84,130,53}
\definecolor{tbs}{RGB}{2,48,71}

\definecolor{prompt}{RGB}{211, 248, 226}
\definecolor{promptable}{RGB}{226, 255, 237}
\definecolor{promptt}{HTML}{06EE62}
\definecolor{repre}{RGB}{228, 193, 249}
\definecolor{repretable}{RGB}{238, 221, 248}
\definecolor{repret}{HTML}{9F00FC}
\definecolor{mixture}{RGB}{237, 231, 177}
\definecolor{mixturetable}{RGB}{250, 246, 210}
\definecolor{mixturet}{HTML}{F1D900}

\title{Continual Learning with Pre-Trained Models: A Survey}

\author{
Da-Wei Zhou$^{1,2}$
\and
Hai-Long Sun$^{1,2}$\and
Jingyi Ning$^1$\and
Han-Jia Ye$^{1,2}$\footnote{Correspondence to: Han-Jia Ye (yehj@lamda.nju.edu.cn)}\And
De-Chuan Zhan$^{1,2}$\\
\affiliations
$^1$National Key Laboratory for Novel Software Technology, Nanjing University\\
$^2$School of Artificial Intelligence, Nanjing University\\
\emails
\{zhoudw, sunhl, yehj, zhandc\}@lamda.nju.edu.cn, ningjy@smail.nju.edu.cn
}

\begin{document}

\maketitle

\input{abstract}
\input{intro}

\input{prelim}

\input{taxonomy}

\input{exp}

\input{future}

{
	\small
	\bibliographystyle{named}
	\bibliography{paper}
}

\end{document}

%% file: abstract.tex
\begin{abstract}
	Nowadays, real-world applications often face streaming data, which requires the learning system to absorb new knowledge as data evolves. Continual Learning (CL) aims to achieve this goal and meanwhile overcome the catastrophic forgetting of former knowledge when learning new ones. 
	Typical CL methods build the model from scratch to grow with incoming data. 
	However, the advent of the pre-trained model (PTM) era has sparked immense research interest, particularly in leveraging PTMs' robust representational capabilities for CL.
	This paper presents a comprehensive survey of the latest advancements in PTM-based CL. We categorize existing methodologies into three distinct groups, providing a comparative analysis of their similarities, differences, and respective advantages and disadvantages.
	Additionally, we offer an empirical study contrasting various state-of-the-art methods to highlight concerns regarding fairness in comparisons.
	The source code to reproduce these evaluations is available at: \url{https://github.com/sun-hailong/LAMDA-PILOT}.
\end{abstract}

%% file: intro.tex
\section{Introduction}

With the rapid development of deep neuron networks, deep learning models have shown promising results in various  applications~\cite{he2015residual}. 
However, the real-world scenario often presents data in a streaming format. Challenges such as privacy concerns and storage limitations prevent the permanent retention of the streaming data, necessitating a learning system capable of continuous adaptation and evolution, a process termed Continual Learning\footnote{Also known as `incremental learning' or `lifelong learning.'} (CL)~\cite{van2022three,de2021survey,masana2022class}. 
A critical issue in CL is the phenomenon of {\em catastrophic forgetting}, where acquiring new knowledge leads to a significant decline in performance on previously learned tasks~\cite{mccloskey1989catastrophic}. Numerous studies have been dedicated to addressing this issue within CL~\cite{gunasekara2023survey,FortinC22,IIM21,sun2023self,wiwatcharakoses2019self,li2022learning}.

Traditionally, CL methods start with models that are ``trained from scratch,'' \ie, beginning with randomly initialized weights. However, the flourishing field of pre-training techniques has opened up new avenues. Utilizing pre-trained models (PTMs), which are developed from extensive datasets and sophisticated techniques~\cite{steiner2021train}, has shown great promise for CL. These PTMs inherently possess a strong {\em generalizability} for a variety of downstream tasks, making PTM-based CL an increasingly popular topic.

\begin{figure}[t]
	\vspace{-4mm}
	\begin{center}
		\includegraphics[width=1\columnwidth]{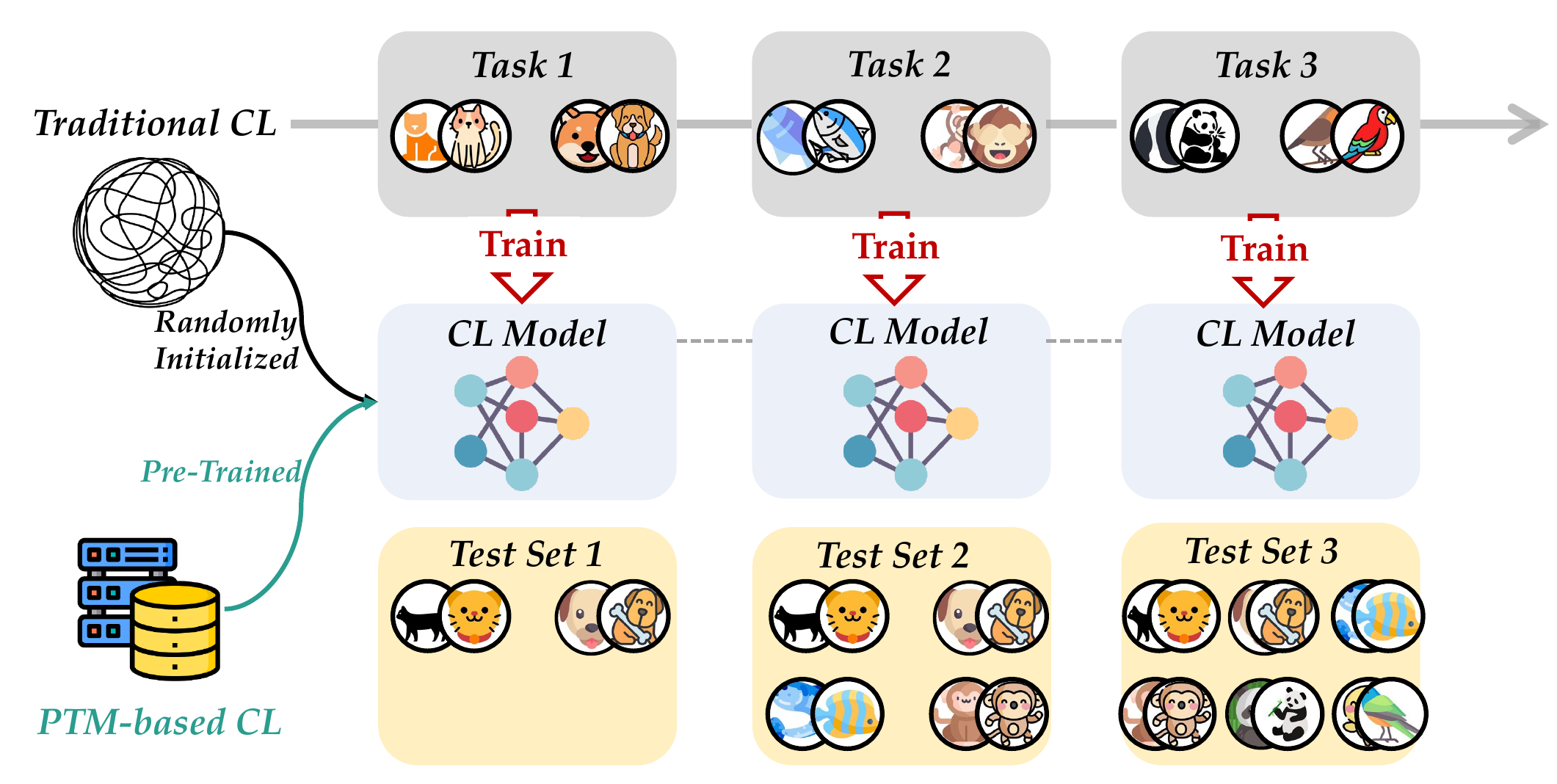}
	\end{center}
	\vspace{-4mm}
	\caption{Illustrations of CL and its variations. 
		New tasks arrive sequentially, and the model needs to learn them incrementally. After the learning process of each task, the model will be evaluated among all seen tasks. Traditional CL methods utilize randomly initialized weights as model initialization, while PTM-based methods make use of substantial and informative data to pre-train the CL model.
	} \label{figure:intro}
		\vspace{-5mm}
\end{figure}

Figure~\ref{figure:intro} illustrates the distinctions between PTM-based and traditional continual learning approaches. Both methodologies employ the CL model within a data stream to adapt to a series of incoming tasks. The objective is for the model to assimilate new information while retaining previously acquired knowledge. This necessitates evaluating the model across all encountered tasks after each new task is learned. The primary divergence between PTM-based and traditional CL lies in the initial setup of the CL model. PTM-based strategies start with a large-scale pre-trained model, whereas traditional methods begin with a model trained from scratch. This difference can be analogized to human learning: traditional methods resemble training an infant to grow up and acquire new knowledge, while PTM-based methods are akin to leveraging the expertise of an adult for the same learning tasks.

In this rapidly evolving field, existing surveys on CL primarily focus on typical algorithms that do not incorporate pre-trained models~\cite{van2022three,de2021survey,masana2022class}. Yet, in the current PTM era, PTM-based CL is emerging as a central area of interest. Observations suggest that the performance of PTM-based CL is approaching the upper bound of continual learning's potential~\cite{zhou2023class}, indicating a promising avenue for practical applications. Consequently, there is an immediate need for a comprehensive, current survey of PTM-based CL to further the advancement of the CL domain.

Specific contributions of our survey are as follows:
	{\bf 1)} We present the first comprehensive survey of recent advancements in pre-trained model-based continual learning, encompassing problem definitions, benchmark datasets, and evaluation protocols. Our systematic categorization of these methods into three subcategories based on their defining characteristics offers a thorough and structured overview of the subject.
{\bf 2)} Our evaluation extends to representative methods in each subcategory across seven benchmark datasets. In addition, we identify a critical factor that can affect the fairness of comparisons in PTM-based continual learning, providing insights into methodological assessments.
	{\bf 3)} We highlight the current challenges and potential future directions in PTM-based continual learning. We intend to shed light on under-researched aspects to spur further investigations that will explore the various possible paths and their interrelations within this field.

%% file: prelim.tex
\section{Preliminaries}

\subsection{Continual Learning}
Continual Learning~\cite{de2021survey} focuses on the learning scenario involving a sequence of tasks $\left\{\D^{1}, \D^{2}, \cdots, \D^{B}\right\}$. The $b$-th dataset $\D^{b}=\left\{\X_b,\Y_b\right\}$ contains the set of input instances and their labels, \ie,  $\X_b=\left\{\x_{i}\right\}_{i=1}^{n_b}$ and $\Y_b=\left\{y_{i}\right\}_{i=1}^{n_b}$.
Among them,   $\x_i \in \R^D$ is an instance of class $y_i \in Y_b$, $Y_b$ is the label space of task $b$.
During the $b$-th training stage, we can only access data from $\D^b$. The goal of continual learning is to continually acquire the knowledge of {\em all seen tasks}, \ie, to fit a model $f(\x)$, and minimize the expected risk:
\begin{equation} \label{eq:totalrisk} 
	f^*=\underset{f \in \mathcal{H}}{\operatorname{argmin}} \enspace \mathbb{E}_{(\mathbf{x}, y) \sim \mathcal{D}_{t}^1\cup\cdots\mathcal{D}_{t}^b} \mathbb{I}(y \neq f(\mathbf{x})) \,.
\end{equation}
In Eq.~\ref{eq:totalrisk}, $\mathcal{H}$ is the hypothesis space, $\mathbb{I}(\cdot)$ is the indicator function which outputs $1$ if the expression holds and $0$ otherwise. $\mathcal{D}_{t}^b$ denotes the data distribution of task $b$. 
Hence, CL models are supposed to work well on all seen tasks, \ie, not only learning new tasks but also not forgetting former ones.

\noindent\textbf{Variations of CL}: There are many specific variations of continual learning based on the definitions of ``tasks''~\cite{van2022three}, \eg, Class-Incremental Learning (CIL), Task-Incremental Learning (TIL), and Domain-Incremental Learning (DIL). Specifically, in the training stage of CIL and TIL, we have $p(\X_b)\neq p(\X_{b^\prime}), \Y_b  \cap \Y_{b^\prime} = \varnothing$ for $b\neq b^\prime$. In other words, the new tasks contain new classes that have not been seen before, and the model is expected to learn new classes while not forgetting the former ones. However, the difference between them lies in the test stage, where TIL provides the task id (\ie, $b$) for the test instance while CIL does not. On the other hand, DIL focuses on the scenario where $p(\X_b)\neq p(\X_{b^\prime}), \Y_b  = \Y_{b^\prime}$ for $b\neq b^\prime$. For example, a new task contains images of the same class but with a domain shift, \eg, cartoon and oil painting.

\subsection{Pre-Trained Models}
Before the prosperity of PTMs, continual learning methods mainly resort to a residual network (\ie, ResNet~\cite{he2015residual}) to serve as the backbone. However, recent years have witnessed the rapid development of transformer-base backbones~\cite{vaswani2017attention}, and most PTM-based CL methods utilize an ImageNet21K~\cite{deng2009imagenet} pre-trained Vision Transformer (ViT)~\cite{dosovitskiy2020image} as embedding function.  Hence, we also focus on ViT as a representative PTM in this paper for its strong representation ability.

Specifically, in ViT, an input image is first divided into non-overlapping patches. These patches are then appended with a class token $\texttt{[CLS]}$ and fed into an embedding layer followed by the vision transformer blocks. We denote the embedded patch features as $\x_e\in\R^{L\times d}$, where $L$ is the length of the sequence and $d$ is the embedding dim. 
In each vision transformer block, there are two main modules, \ie, a multi-head self-attention layer (MSA) and a two-layer MLP. 
The patch features are forwarded by $N$ cascaded transformer blocks, and we utilize the final $\texttt{[CLS]}$ token as the feature for recognition. 
In the following discussions, we assume the availability of a pre-trained ViT on ImageNet as the initialization of $f(\x)$. We decompose the classification model into two parts: $f(\x)=W^{\top}\phi(\x)$, where $\phi(\cdot):\mathbb{R}^{D} \rightarrow \mathbb{R}^{d}$ is the embedding function (\ie, the embedded $\texttt{[CLS]}$ token) and $W\in\mathbb{R}^{d\times |\bigcup_{b=1}^B{Y}_{b}|}$ is the classification head.

\subsection{New Insights in Continual Learning Brought by Pre-Trained Models}

Compared to training the embedding function from scratch, utilizing pre-trained models brings two major characteristics. Firstly, PTMs are born with ``{\em generalizability}'' compared to a randomly initialized model. From the representation learning perspective, the ultimate goal of continual learning is to learn a suitable embedding to capture all seen tasks, while PTMs provide a strong and generalizable feature extractor in the beginning. Hence, algorithms can be designed upon the frozen backbone in a non-continual manner~\cite{zhou2023revisiting}. 

On the other hand, the structure of ViTs enables lightweight tuning with frozen pre-trained weights. Techniques like visual prompt tuning~\cite{jia2022visual} and adapter learning~\cite{chen2022adaptformer} enable quick adaptation of PTMs to the downstream task while preserving generalizability. Hence, continual learning with PTMs shows stronger performance in resisting forgetting than training from scratch~\cite{cao2023retentive}.

%% file: taxonomy.tex
\section{Continual Learning with PTMs}

\begin{figure}[t]
	\vspace{-4mm}
	\begin{center}
		\begin{overpic}[width=1\columnwidth]{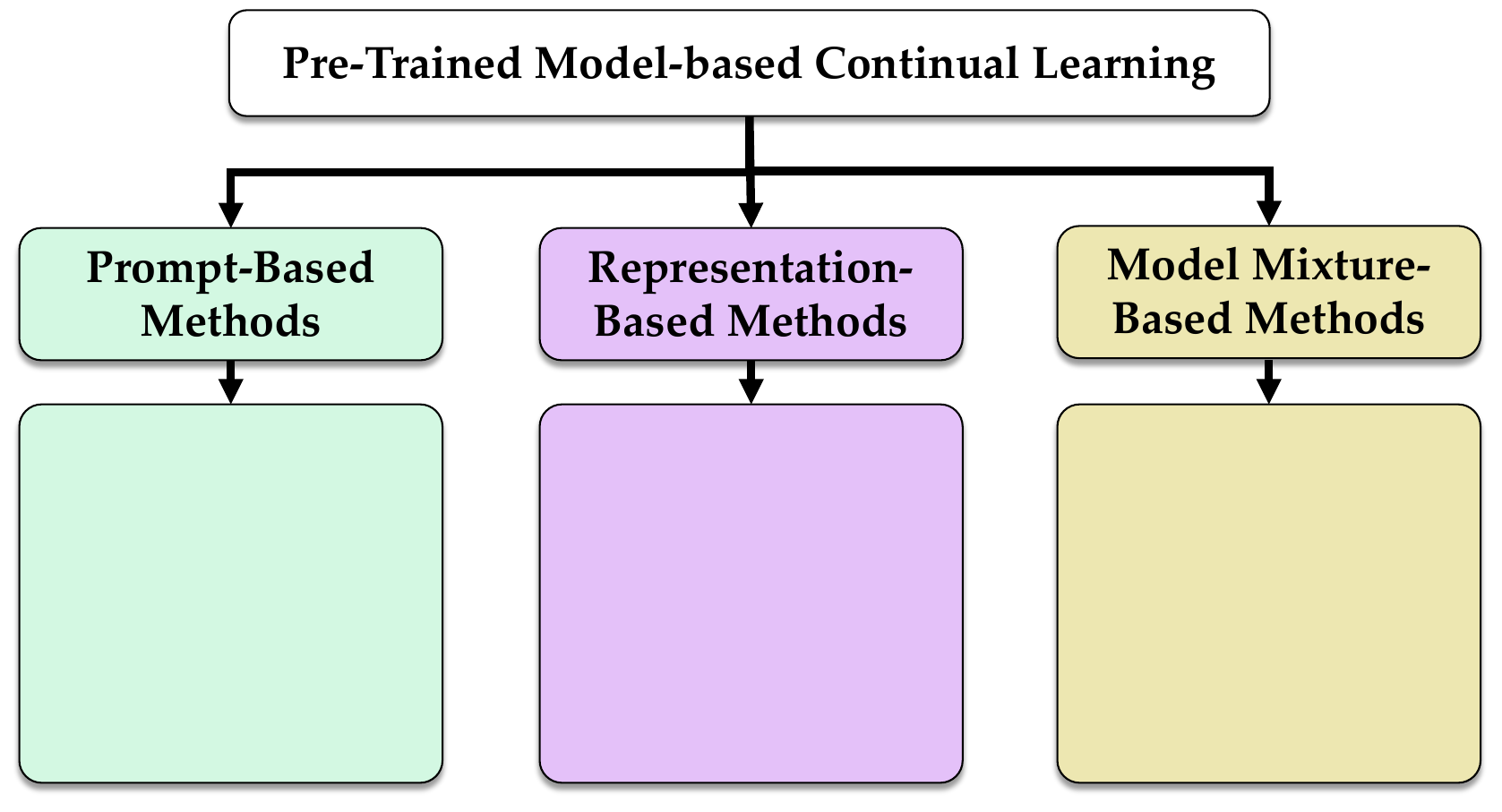}
			\put(5,24){\tiny {\cite{wang2022learning}}}
			\put(5,20.5){\tiny {\cite{wang2022dualprompt}}}
			\put(5,17){\tiny {\cite{wang2022s}}}
			\put(5,13.5){\tiny {\cite{smith2023coda}}}
			\put(5,10){\tiny {\cite{jung2023generating}}}
			\put(5,6.5){\tiny {\cite{tang2023prompt}}}
			\put(5,3){\tiny {\cite{khan2023introducing}}}

			\put(39,24){\tiny {\cite{zhou2023revisiting}}}
			\put(39,20.5){\tiny {\cite{zhou2024expandable}}}
			\put(39,17){\tiny {\cite{mcdonnell2023ranpac}}}
			\put(39,13.5){\tiny {\cite{ahrens2023read}}}
			\put(39,10){\tiny {\cite{Zhang_2023_ICCV}}}
			\put(39,6.5){\tiny {\cite{zheng2023learn}}}
			\put(39,3){\tiny {\cite{guo2024federated}}}
			
			\put(75,24){\tiny {\cite{zhou2023learning}}}
			\put(75,20.5){\tiny {\cite{Zheng_2023_ICCV}}}
			\put(75,17){\tiny {\cite{marouf2023weighted}}}
			\put(75,13.5){\tiny {\cite{wang2023isolation}}}
			\put(75,10){\tiny {\cite{chen2023promptfusion}}}
			\put(75,6.5){\tiny {\cite{Gao_2023_ICCV}}}
			\put(75,3){\tiny {\cite{wang2023hierarchical}}}

		\end{overpic}
	\end{center}
	\vspace{-4mm}
	\caption{ Taxonomy of PTM-based CL. We classify them into three subcategories, \ie, prompt-based ({\color{prompt}$\blacksquare$}), representation-based ({\color{repre}$\blacksquare$}), and model mixture-based ({\color{mixture}$\blacksquare$}).   Different colors  indicate different categories, and we list representative works in the boxes.
	} \label{figure:taxonomy}
	\vspace{-4mm}
\end{figure}
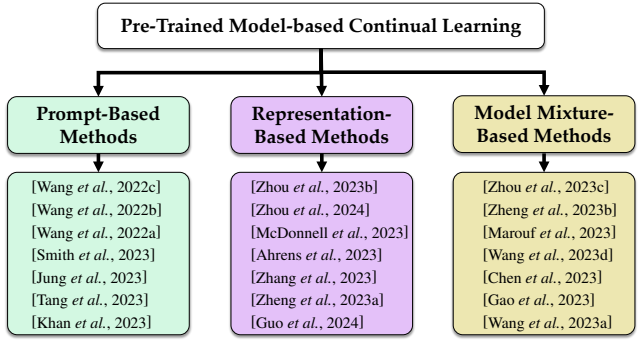

We taxonomize current PTM-based CL studies into three categories based on their different ideas to tackle the learning problem, \ie, prompt-based methods, representation-based methods, and model mixture-based methods. These categories utilize different aspects of pre-trained models to facilitate continual learning. For example, given the strong generalization ability of PTMs, prompt-based methods resort to prompt tuning~\cite{jia2022visual} to exert lightweight updating of the PTM. Since the pre-trained weights are kept unchanged, the generalizability of PTMs can be preserved, and forgetting is thus alleviated. Similarly, representation-based methods directly utilize the generalizability of PTMs to construct the classifier. Lastly, model mixture-based methods design a set of models in the learning process and utilize model merging, model ensemble, and other mixture techniques to derive a final prediction. We show the taxonomy of PTM-based CL and list representative works in Figure~\ref{figure:taxonomy}. In the following section, we introduce each category and discuss their pros and cons in depth.

\subsection{Prompt-based Methods}

Observing the strong generalization ability of PTMs, how to tune the PTM leads to a trade-off ---
fully finetuning the weights to capture downstream tasks will erase the generalizable features, while fixing the backbone cannot encode downstream information into the backbone. To this end, visual prompt tuning (VPT)~\cite{jia2022visual} reveals a promising way to utilize lightweight trainable modules, \eg, prompts, to adjust the PTM. Specifically, it prepends a set of learnable parameters $P\in\R^{p\times d}$ (\ie, prompts) to the patch features $\x_e$. Hence, the model treats the concatenation of $[P, \x_e]$ as the input of the vision transformer blocks and minimizes the cross-entropy loss to encode task-specific information into these prompts with pre-trained weights frozen:
\begin{equation}  \label{eq:prompt}
	\min_{P\cup W} \sum_{(\x,y)\in \D^b} \ell \left(W^\top
	\phi\left(\x; P \right)	, y\right) \,,
\end{equation}
where $\phi\left(\x; P \right)$ represents the prompted features by prepending the prompts. Optimizing Eq.~\ref{eq:prompt} enables the model to encode task-specific information (\ie, the crucial features for $\D^b$) into the prompts. Hence, many works are designed to utilize prompt tuning for CL.

\noindent\textbf{Prompt Pool}: Although Eq.~\ref{eq:prompt} enables the lightweight tuning of a pre-trained model, sequentially optimizing a single prompt with new tasks will suffer catastrophic forgetting, \ie, overwriting the prompt weights of former tasks leads to the incompatible representations between former tasks and latter ones. Hence, many works~\cite{wang2022learning,wang2022dualprompt,smith2023coda} propose to design the prompt pool, which collects a set of prompts, \ie, $\mathbf{P}=\{P_1, P_2, \cdots, P_M\}$, where $M$ is the size of the pool. The prompt pool can be seen as the external memory of the CL model, enabling instance-specific prompting during training and inference. Hence, the forgetting of a single prompt can be alleviated, while it requires a proper {\em prompt selection} mechanism.

\noindent\textbf{Prompt Selection}: With a set of prompts, we need to decide which prompt(s) to use for the specific instance, \ie, to define a retrieval function $g(\x)$ that selects instance-specific prompts. Prompt retrieval becomes the core problem in prompt-based methods, and many works design different variations. L2P~\cite{wang2022learning} designs a key-query matching strategy, which assigns a learnable key $\mathbf{k}\in\R^d$ to each prompt. In this case, the prompt pool is formulated as $\mathbf{P}=\{(\mathbf{k}_1, P_1),(\mathbf{k}_2, P_2),\cdots,(\mathbf{k}_M, P_M)\}$. To retrieve instance-specific prompts, it utilizes a PTM without prompting (\ie, $\phi(\cdot)$) to encode the features into the key's embedding space and select prompts with similar keys:
\begin{equation} \label{eq:l2p}
	\mathbf{K}_{\x}=\underset{\left\{s_i\right\}_{i=1}^N \subseteq[1, M]}{\operatorname{argmin}} \quad \sum_{i=1}^N \gamma\left(\phi({\x}), \mathbf{k}_{s_i}\right) \,,
\end{equation}
where $\left\{s_i\right\}_{i=1}^N$ is the selected index set and $\mathbf{K}_{\x}$ is the selected top-N keys. $\gamma(\cdot, \cdot)$ denotes the cosine distance. Eq.~\ref{eq:l2p} selects the most similar keys to the query instance, and the model optimizes the corresponding values (\ie, prompts) during the learning process: 
\begin{equation}  \label{eq:l2p-optimize}
	\min_{\mathbf{K}\cup\mathbf{P}\cup W} \sum_{(\x,y)\in \D^b} \ell \left(W^\top
	\phi\left(\x; \mathbf{P}_\x \right)	, y\right) + \lambda \sum_{\mathbf{K}_{\x}} \gamma\left(\phi(\x), \mathbf{k}_{s_i}\right)\,.
\end{equation}
Hence, optimizing Eq.~\ref{eq:l2p-optimize} also forces the keys to be similar to the encoded features. The above query-key matching process is an Expectation-Maximization (EM) procedure~\cite{moon1996expectation,yadav2023exploring}. Specifically, in the E-step, the top-N keys are selected based on their similarity to the query feature. In the M-step, the keys are then pulled closer to the query. 

Motivated by L2P, many works are proposed to improve the selection process. DualPrompt~\cite{wang2022dualprompt} explores the significance of prompt depth by attaching prompts to different layers. It also decouples the prompts into general and expert ones. Among them, general prompts are designed to encode the task-generic information, which is shared among all tasks. By contrast, expert prompts are task-specific, and the number is equal to that of tasks. It utilizes the same retrieval strategy in Eq.~\ref{eq:l2p} during inference. PP-TF~\cite{yadav2023exploring} applies a similar strategy in code generation models. S-Prompt~\cite{wang2022s} also considers a task-specific prompt strategy, which expands the prompt pool with a new prompt when learning a new task. Instead of key-query matching, it builds the task centers by conducting K-means clustering in every task and utilizes a KNN search to find the most similar task to get the prompt. MoP-CLIP~\cite{nicolas2024mop} extends S-Prompt by combining multiple prompts during inference.

\noindent\textbf{Prompt Combination}: While selecting prompts from the prompt pool sounds reasonable, the matching process in Eq.~\ref{eq:l2p} is still a hard matching that can reproduce limited choices. Correspondingly, CODA-Prompt\cite{smith2023coda} suggests building an attention-based prompt from the prompt pool. During prompt retrieval, it utilizes the query feature $\phi(\x)$ to calculate an attention vector to all keys and utilize the attention results to create a weighted summation over the prompt components:
\begin{equation}  \label{eq:coda} \textstyle
	P=\sum_{m=1}^M \gamma(\phi(\x) \odot \mathbf{a}_m, \mathbf{k}_m) P_m \,, 
\end{equation}
where $\mathbf{a}_m \in\R^d$ is the learnable attention vectors of the corresponding prompt, and $\odot$ denotes Hadamard product. Eq.~\ref{eq:coda} calculates the attention score between the input feature and the prompt keys via element-wise multiplication. Hence, if the query instance is more similar to a key vector, the corresponding prompt value will play a more important role in the final constructed prompt. Since it treats the prompts like `bases' in the prompt space, it also designs an extra orthogonality loss to enhance prompt diversity.

\begin{figure}[t]
	\vspace{-4mm}
	\begin{center}
		\includegraphics[width=1\columnwidth]{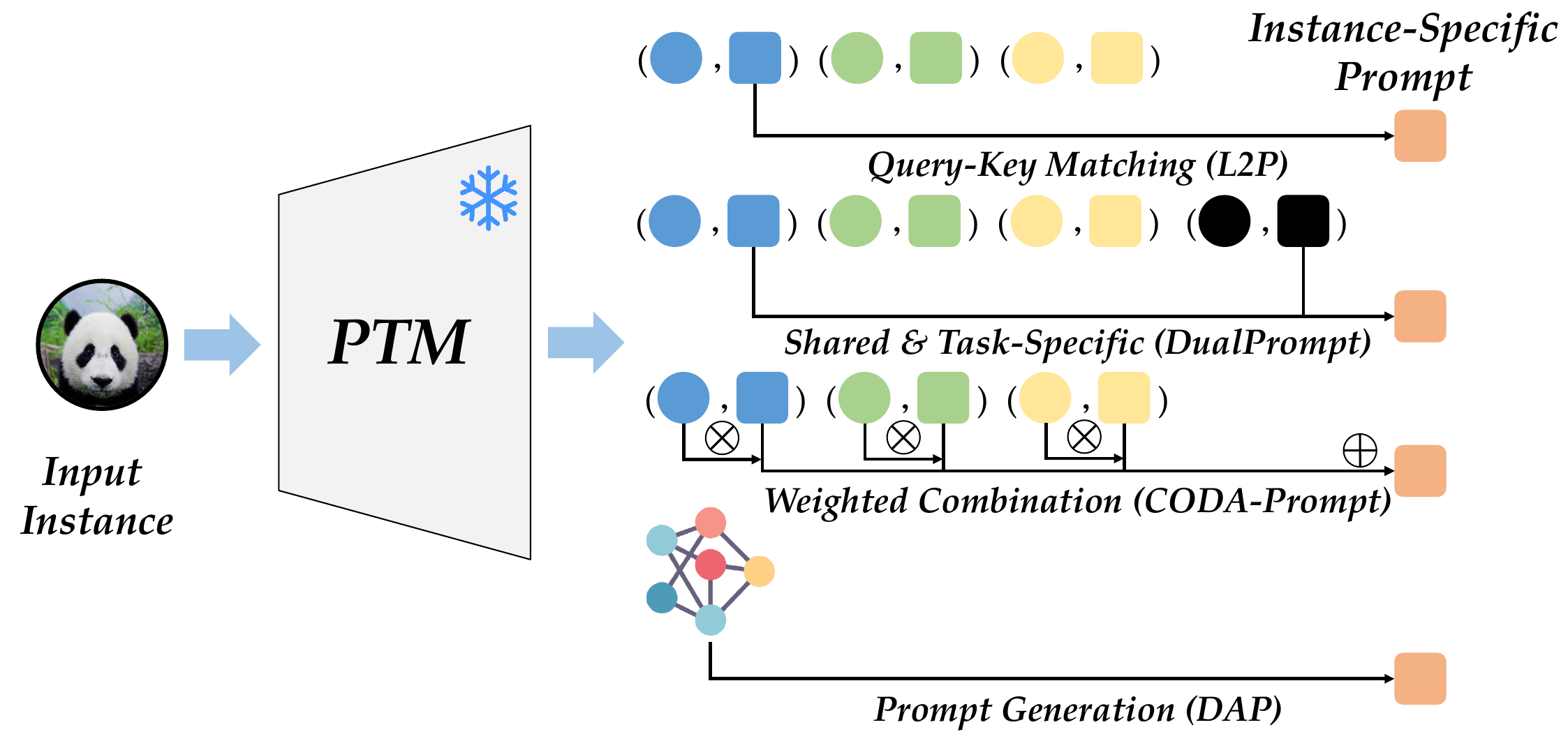}
	\end{center}
		\vspace{-5mm}
	\caption{ Different kinds of prompt selection, including key-value matching, shared and task-specific retrieval, attention-based combination, and instance-specific prompt generation. 
	} 
	\label{figure:prompt}
		\vspace{-4mm}
\end{figure}

\noindent\textbf{Prompt Generation}: While CODA-Prompt addresses the attention-based prompt combination, the combination process is still restricted by the prompt pool. Hence, many works move further to design meta-networks that can {\em generate} instance-specific prompts. Correspondingly, DAP~\cite{jung2023generating} achieves this goal by encoding prompt generation into an MLP network. It generates instance-specific prompts via:
\begin{equation} \label{eq:dap}
		P =\left(\gamma_e \operatorname{MLP}\left(\operatorname{LN}(\phi(\x))^{\top}\right)+\beta_e\right)^{\top} \,,
\end{equation}
where LN denotes layer normalization, $\gamma_e$, and $\beta_e$ are produced by linear transformations of the task prediction, serving as the weight and bias in prompt generation. Unlike the input-level prompt generation in Eq.~\ref{eq:dap}, APG~\cite{tang2023prompt} utilizes the attention mechanism for prompt generation at the middle layers of ViT.

\noindent\textbf{Summary of Prompt-based Methods}: We summarize the way of prompt selection in Figure~\ref{figure:prompt}, including the way of prompt retrieval in L2P, task-specific and general prompts in DualPrompt, attention-based combination in CODA-Prompt and prompt generation in DAP. Instead of selecting prompts, several works~\cite{liu2022incremental,razdaibiedina2022progressive} also consider appending all prompts to the query instance or learning visual prompts (\ie, pixel-level parameters)~\cite{liu2023dual,gan2023decorate}. Apart from the single visual modality, the model can also utilize textual information~\cite{radford2021learning} in learning and selecting suitable prompts with pre-trained vision-language models~\cite{khan2023introducing,villa2023pivot,wang2023attriclip,khattak2023self}.

\noindent\textbf{Pros \& Cons}: Prompt-based methods strike a balance between pre-trained knowledge and downstream tasks with lightweight prompts, yielding many advantages. Firstly, prompts help bridge the domain gap and effectively encode task-specific knowledge. Secondly, since these lightweight modules have the same dimension as the features, saving prompts is parameter-efficient, which is naturally suitable for some edge scenarios like federated learning~\cite{guo2024federated}. Lastly, learning the prompt pool acts as the external memory of the PTM, enabling adaptive knowledge retrieval and instance-specific prediction.

However, there are also some drawbacks. Firstly, some works~\cite{moon2023online} find the prompt selection process in Eq.~\ref{eq:l2p} converges to a single point, making the prompt selection only concentrate on the specific subset. Besides, since the keys and prompt values keep changing throughout the learning process, the updating of these parameters will erase that of former tasks. 
This further leads to matching-level and prompt-level forgetting, making the prompt selection process become the bottleneck in continual learning.
Furthermore, if we use a fixed-size prompt pool, the representation ability will be restricted. By contrast, if the prompt pool grows as data evolves, it will result in a mismatch between training and testing since new prompts may be retrieved for old tasks. Lastly, although prompt-based methods reveal a promising solution for PTM-based CL, some works~\cite{zhou2023revisiting} find their performance lower than a simple prototype-based baseline (as discussed in Section~\ref{sec:representation-based}). While some prompt-based methods~\cite{jung2023generating} show surprising results, there are some concerns about the comparison fairness due to the batch-wise prompt section (as discussed in Section~\ref{sec:exp}).

\subsection{Representation-based Methods} \label{sec:representation-based}

Observing the strong representation ability of PTMs, one may wonder if they have already mastered the knowledge to classify new tasks. In other words, how can we measure the inherent ability of PTMs on these downstream tasks? Borrowing the idea from representation learning, SimpleCIL~\cite{zhou2023revisiting} suggests a simple way to achieve this goal. Facing the continual data stream, it freezes the pre-trained weights and extracts the center $\cc$ (\ie, prototype) of each class:
\begin{align} \label{eq:prototype}\textstyle
	\cc_i=\frac{1}{K}{\sum_{j=1}^{|\mathcal{{D}}^b|}\mathbb{I}(y_j=i)\phi(\x_j)}
	\,,
\end{align}
where $K={\sum_{j=1}^{|\mathcal{{D}}^b|}\mathbb{I}(y_j=i)}$. In Eq.~\ref{eq:prototype}, the embeddings of the same class are averaged, leading to the most common pattern of the corresponding class. Hence, it can serve as the `classification criterion' or `template'~\cite{van2022three} during inference. Correspondingly, SimpleCIL directly replaces the classifier weight of the $i$-th class with prototype ($\w_i=\cc_i$), and utilize a cosine classifier for classification, \ie, $f(\mathbf{x})=\frac{W^\top\phi(\x)}{\|W\|_{2}\|\phi(\x)\|_{2}}$. Hence, facing a new task, we can calculate and replace the classifier for each class with the embedding frozen. Surprisingly, this simple solution shows superior performance than many prompt-based methods, \eg, L2P and DualPrompt. It indicates that PTMs already possess generalizable representations, which can be directly utilized for downstream tasks. A similar phenomenon has also been found in~\cite{janson2022simple}, and \cite{zheng2023learn} applies it to large language models.

\noindent\textbf{Concatenating Backbones}: Observing the strong generalizability of PTMs, ADAM~\cite{zhou2023revisiting} moves a step further by comparing the performance of new classes between the prototype-based classifier and fully-finetuned model. Surprisingly, it finds PTMs can achieve better performance on new classes if adapted to the downstream tasks. It indicates that PTMs, although generalizable, do not possess the task-specific information for the downstream data. Hence, ADAM suggests finetuning the PTM with parameter-efficient modules (\eg, prompts~\cite{jia2022visual} or adapters~\cite{chen2022adaptformer}) and concatenating features of the pre-trained and adapted models:
 \begin{align} \label{eq:prototype-adam}\textstyle
	\cc_i=\frac{1}{K}
	{\sum_{j=1}^{|\mathcal{{D}}^b|}\mathbb{I}(y_j=i)[\phi(\x_j), \phi(\x_j; \text{PEFT})]}
	\,,
\end{align}
where $\phi(\x_j; \text{PEFT})$ indicates the finetuned model. In Eq.~\ref{eq:prototype-adam}, the adaptation process bridges the domain gap between pre-trained and downstream datasets, and the concatenated features possess generalized (\ie, PTM) and task-specific (\ie, finetuned model) information. Hence, ADAM further improves the performance compared to SimpleCIL. Recently, EASE~\cite{zhou2024expandable} further concatenates the feature representations of multiple task-specific backbones, leading to state-of-the-art performance. It designs a semantic mapping strategy for classifier complement to compensate for the ever-expanding features and the previous classifiers.

\noindent\textbf{Utilizing Random Projection}: Based on ADAM, RanPAC~\cite{mcdonnell2023ranpac} further finds that prototypes calculated by Eq.~\ref{eq:prototype-adam} often correlate between classes. Hence, it suggests using an online LDA classifier to remove class-wise correlations for better separability. Furthermore, to make the feature distribution for a Gaussian fit, it designs an extra random projection layer $\text{Proj}(\cdot):\mathbb{R}^{d} \rightarrow \mathbb{R}^{K}, K\gg d$ to project features into the high dimensional space. Afterward, prototypes in the projected space are calculated, \ie, $\cc_i=\frac{1}{K}{\sum_{j=1}^{|\mathcal{{D}}^b|}\mathbb{I}(y_j=i)\text{Proj}(\phi(\x_j; \text{PEFT})})$. Moreover, LayUP~\cite{ahrens2023read} further finds the strong representation ability also lies in other deep layers of the transformer block. It treats the concatenation of the last $k$ layer features as the representation and trains an online LDA based on it.

\noindent\textbf{Slow Learner with Feature Replay}: In Eq.~\ref{eq:prototype-adam}, the model's generalizability and adaptivity are maintained by backbone concatenation. By contrast, there are also works aiming for the intersection between pre-trained and fully-adapted models. SLCA~\cite{Zhang_2023_ICCV} suggests tuning the embedding function $\phi$ with a small learning rate while tuning the classifier $W$ with a large learning rate. This enables a gradual fitting of the features and quick adaptation of the classifiers. To resist forgetting former classifiers, it follows~\cite{zhu2021prototype} to model class-wise feature distributions and replay them to calibrate the classifier.

\noindent\textbf{Pros \& Cons}: Representation-based methods aim to take full advantage of the pre-trained features, which show competitive performance in various tasks. This line of work has many advantages. Firstly, since class prototypes represent the most common pattern of the corresponding class, building recognition models with them is intuitive and interpretable. 
Utilizing a prototype-based classifier also provides a simple yet effective way to investigate the `baseline' of PTM-based CL.
Furthermore, this line of work mainly freezes the backbone and updates the classifier weight. The lightweight update cost makes them feasible in real-world applications, \eg, \cite{guo2024federated} applies similar tricks to federated learning by synchronizing global prototypes in various clients.

However, there are also some drawbacks. Firstly, concatenating features from different models to formulate the class prototype ignores the redundancy across models. For example, the shared features could be extracted repeatedly in different backbones without a pruning strategy. Secondly, when the downstream task involves multiple domains, adapting the model within the first stage (as in Eq.~\ref{eq:prototype-adam}) is insufficient to bridge the domain gap across datasets. In that case, continually adjusting the backbone could be more suitable to extract task-specific features.

\subsection{Model Mixture-based Methods}

The challenge of continual learning has been alleviated with the help of PTMs, enabling CL algorithms to start from a provident starting point.
Hence, adapting PTMs to downstream tasks becomes simple, while preventing forgetting in the adaptation process requires more attention. To this end, model mixture-based methods aim to create a set of models during the continual learning process and conduct model ensemble or model merge during inference. Since information from multiple stages is mixed for the final prediction, catastrophic forgetting can be thus alleviated.

\noindent\textbf{Model Ensemble}: Since PTMs show generalizable features, creating a set of models based on the PTM becomes possible. ESN~\cite{wang2023isolation} creates a set of classifiers individually based on the same PTM during the learning process, \ie, it initializes and trains a new classifier head when facing a new task. During inference, it designs a voting strategy for these classifier heads by adopting a set of temperatures. LAE~\cite{Gao_2023_ICCV} adopts a similar inference strategy by choosing the max logit across different models.

Since the core factor in an ensemble depends on the {\em variance} of learners, several works aim to enhance the diversity among models instead of building a set of classifiers with the same PTM. PromptFusion~\cite{chen2023promptfusion} utilizes a pre-trained ViT and a CLIP~\cite{radford2021learning} and dynamically combines the logits during inference, \ie, $f(\x)=\lambda f_{\text{ViT}}(\x)+(1-\lambda)f_{\text{CLIP}}(\x)$. 
Different from the ensemble of multiple backbones, PROOF~\cite{zhou2023learning} designs a more comprehensive inference format with only a single CLIP. Since CLIP enables cross-modal matching for visual and textual features, PROOF designs a three-level ensemble considering image-to-text, image-to-image prototype, and image-to-adjusted text with cross-modal fusion.

\begin{table*}[t]
	\vspace{-5mm}
	\caption{ Average and last continual learning performance on seven datasets.  `IN-R/A' stands for `ImageNet-R/A,' `ObjNet' stands for `ObjectNet,' and `OmniBench' stands for `OmniBenchmark.' 
		Different colors ({\color{prompt}$\blacksquare$} {\color{repre}$\blacksquare$} {\color{mixture}$\blacksquare$}) indicate different categories of methods. `DAP w/o BI' indicates DAP without batch information.
	}\label{tab:benchmark}
	\vspace{-3mm}
	\centering
	\resizebox{1.0\textwidth}{!}{%
		\begin{tabular}{@{}lccccccccc cccccccc}
			\toprule
			\multicolumn{1}{l}{\multirow{2}{*}{Method}} & 
			\multicolumn{2}{c}{CIFAR B0 Inc5} & \multicolumn{2}{c}{CUB B0 Inc10} 
			& \multicolumn{2}{c}{IN-R B0 Inc5}
			& \multicolumn{2}{c}{IN-A B0 Inc20}
			& \multicolumn{2}{c}{ObjNet B0 Inc10}
			& \multicolumn{2}{c}{OmniBench B0 Inc30}
			& \multicolumn{2}{c}{VTAB B0 Inc10} \\
			& {$\bar{\mathcal{A}}$} & ${\mathcal{A}_B}$  
			& {$\bar{\mathcal{A}}$} & ${\mathcal{A}_B}$
			& {$\bar{\mathcal{A}}$} & ${\mathcal{A}_B}$   
			& {$\bar{\mathcal{A}}$} & ${\mathcal{A}_B}$ 
			& {$\bar{\mathcal{A}}$} & ${\mathcal{A}_B}$ 
			& {$\bar{\mathcal{A}}$} & ${\mathcal{A}_B}$ 
			& {$\bar{\mathcal{A}}$} & ${\mathcal{A}_B}$ 
			\\
			\midrule
			\rowcolor{promptable}	L2P   & 85.94 & 79.93 &67.05 & 56.25 & 66.53 & 59.22 &  49.39 & 41.71 &  63.78 & 52.19 &73.36 & 64.69 & 77.11 & 77.10\\
			\rowcolor{promptable}	DualPrompt    &87.87 & 81.15& 77.47 & 66.54 & 63.31 & 55.22 & 53.71 & 41.67 & 59.27 & 49.33 & 73.92 & 65.52 & 83.36 & 81.23\\
			\rowcolor{promptable}	CODA-Prompt & 89.11 & 81.96 & 84.00 & 73.37 & 64.42 &55.08 & 53.54 & 42.73 & 66.07 &53.29 &77.03 &68.09 &83.90 &83.02\\
			\rowcolor{promptable}	DAP  & 94.54 & 90.62 & 94.76 & 94.63& 80.61 &74.76&54.39 &46.32&72.08 &59.51& 86.44 &80.65 &84.65& 84.64\\
			\rowcolor{promptable}	DAP w/o BI   &68.07 &58.16&65.27 &52.05&50.40 &37.99&34.48 &21.84&50.47 &37.55&65.43 &52.53&79.63 &79.87\\
			\rowcolor{repretable}	SimpleCIL   &  87.57 & 81.26 & 92.20 & 86.73 & 62.58 & 54.55 & 59.77 & 48.91 & 65.45 & 53.59 & 79.34 & 73.15 & 85.99 & 84.38\\
			\rowcolor{repretable}	ADAM + VPT-D & 88.46 & 82.17 & 91.02 &84.99 & 68.79 & 60.48 & 58.48 & 48.52 & 67.83 & 54.65 &  81.05 &  74.47 & 86.59 & 83.06\\
			\rowcolor{repretable}	ADAM + SSF  & 87.78 & 81.98   & 91.72 &86.13& 68.94 & 60.60 &61.30 & 50.03 & 69.15 & 56.64 &  80.53 & 74.00 & 85.66 & 81.92\\
			\rowcolor{repretable}	ADAM + Adapter &   90.65 &  85.15 &92.21 &86.73 & 72.35 & 64.33 & 60.47 &49.37 &  67.18 & 55.24 &  80.75 & 74.37 &  85.95 & 84.35\\
			\rowcolor{repretable}	RanPAC  & 93.51 & 89.30& 93.13 & 89.40& 75.74 & 68.75& 64.16 & 52.86& 71.67 & 60.08& 85.95 & 79.55&92.56 &91.83\\
			\rowcolor{repretable} EASE &  91.51 &  85.80 & 92.23& 86.81 & 78.31& 70.58 & 65.34 & 55.04 & 70.84 & 57.86 & 81.11& 74.85& 93.61& 93.55\\
			\rowcolor{mixturetable}	HiDe-Prompt & 91.22 & 89.92 & 89.75 & 89.46 & 76.20 & 74.56 & 61.41 & 49.27 & 70.13 & 62.84 & 76.60 & 77.01 & 91.24 & 92.78\\
			\rowcolor{mixturetable}	ESN & 87.15 & 80.37& 65.69 & 63.10& 60.69 & 55.13& 44.06 & 31.07& 63.73 & 52.55& 75.32 & 66.57& 81.52 & 62.15\\
			\bottomrule
		\end{tabular}	}
	\vspace{-3mm}
\end{table*}

\noindent\textbf{Model Merge}: Another line of work considers model merge, which combines multiple distinct models into a single unified model without requiring additional training. LAE~\cite{Gao_2023_ICCV} defines the online and offline learning protocol, where the online model is updated with cross-entropy loss, aiming to acquire new knowledge in new tasks. By contrast, the offline model is updated via model merge, \eg, Exponential Moving Average (EMA):
\begin{equation} \label{eq:lae}
	\boldsymbol{\theta}^{\text {Offline}} \leftarrow \alpha \cdot \boldsymbol{\theta}^{\text {Offline}}+(1-\alpha) \cdot \boldsymbol{\theta}^{\text {Online}} \,,
\end{equation}
where $\alpha$ is the trade-off parameter. Notably, LAE only applies Eq.~\ref{eq:lae} to the parameter-efficient tuning modules (\eg, prompt). It utilizes the max logit of online and offline models for inference.
Hide-Prompt~\cite{wang2023hierarchical} also applies a similar prompt merge after each continual learning stage. 

Like LAE, ZSCL~\cite{Zheng_2023_ICCV} applies the merging technique to the CLIP model, aiming to maintain its zero-shot performance during continual learning. However, it finds that the performance is not robust with the change of the trade-off parameter in Eq.~\ref{eq:lae}. Hence, it proposes to merge the parameters every several iterations, enabling the creation of a smooth loss trajectory during model training. Moreover, noticing that Eq.~\ref{eq:lae} assigns equal importance to each parameter during merging, CoFiMA~\cite{marouf2023weighted} argues different parameters shall have different importance to the task. Hence, it inserts Fisher information as the estimated importance of each parameter during the merging process.

\noindent\textbf{Pros \& Cons}: In PTM-based CL, building multiple models for mixture upon the pre-trained weights is intuitive. Hence, there are some advantages of model mixture-based methods. Firstly, learning multiple models enables a diverse decision within model sets. Consequently, using model merging or ensemble leads to naturally more robust results. Secondly, since models are directly merged for a unified prediction, the weight of former and latter models can be adjusted to highlight the importance of knowledge  shared among different stages. Lastly, since the set of models will be merged during inference, the final inference cost will not increase as more models are added to the model set. Re-parameterization techniques can also be applied for model merge, enabling a restricted model size for edge devices~\cite{zhou2023learning,wang2023orthogonal}.
 
However, we also notice some drawbacks of model mixture-based methods. Firstly, designing a model ensemble requires saving all historical models and consuming a large memory buffer. While model merge-based methods do not require such a large cost, merging the weights of the large backbone also requires many extra computations. Secondly, deciding which parameters to merge remains an open problem, making the merging solution heuristic and hand-crafted.

%% file: exp.tex
\section{Experiments} \label{sec:exp}

\noindent{\bf Datasets}: Since PTMs are often trained with ImageNet21K~\cite{deng2009imagenet}, evaluating methods with ImageNet is meaningless. Consequently, we follow~\cite{zhou2023revisiting,mcdonnell2023ranpac} to evaluate the performance on CIFAR100~\cite{krizhevsky2009learning}, CUB200~\cite{WahCUB2002011}, ImageNet-R~\cite{hendrycks2021many}, ImageNet-A~\cite{hendrycks2021natural}, ObjectNet~\cite{barbu2019objectnet}, Omnibenchmark~\cite{zhang2022benchmarking} and VTAB~\cite{zhai2019large}.
Apart from typical benchmarks for CL (\eg, CIFAR and CUB), the other five datasets are acknowledged to have {\em large domain gap} with ImageNet, making the PTM less generalizable and increasing the difficulty of CL.

\noindent {\bf Dataset split:} Following~\cite{zhou2023class}, we denote the data split as `B-$m$, Inc-$n$,' \ie, the first dataset contains $m$ classes, and each following dataset contains $n$ classes. $m=0$ means the total classes are equally divided into each task. 
Before splitting, we randomly shuffle all classes with the \emph{same} random seed~\cite{zhou2023class} for a fair comparison.

\noindent {\bf Training details:}
We use PyTorch to deploy all models with the {\em same} network backbone.
We follow~\cite{wang2022learning} to choose the most representative ViT pre-trained on ImageNet21K, \ie,  ViT-B/16-IN21K.

\noindent {\bf Performance measure:} Denote the Top-1 accuracy after the $b$-th stage as $\mathcal{A}_b$, we follow~\cite{zhou2023class} to use 
 $\mathcal{A}_B$ (last stage accuracy) and $\bar{\mathcal{A}}=\frac{1}{B}\sum_{b=1}^{B}\mathcal{A}_b$ (average performance along incremental stages) as performance measures.

\noindent\textbf{Experimental results:} Following the taxonomy in Figure~\ref{figure:taxonomy}, we compare nine methods from the three categories. Among them,  L2P, DualPrompt, CODA-Prompt, and DAP are {\color{promptt}{\bf prompt-based}} methods; SimpleCIL, ADAM, RanPAC, and EASE are {\color{repret}{\bf representation-based}} methods; ESN and HiDe-Prompt are  {\color{mixturet}{\bf model mixture-based}} methods. We report the results on seven benchmark datasets in Table~\ref{tab:benchmark} and use different colors to represent methods of different categories. From these results, we have three main conclusions. 
{\bf 1)} Almost all methods perform well on typical CL benchmarks, \ie, CIFAR100, while some of them have problems with benchmarks that have large domain gaps to the pre-trained dataset (\eg, ImageNet-A). This indicates that more challenging benchmarks should be raised to serve as the CL benchmarks in the era of PTMs.
{\bf 2)} We observe that representation-based methods (\eg, ADAM and RanPAC) show more competitive performance than the others (except for DAP, which will be discussed later). This indicates that representation in prompt-based and model mixture-based methods can be further cultivated to improve their performance. {\bf 3)} We observe that the simple baseline SimpleCIL shows better performance than typical prompt-based methods (\eg, L2P and DualPrompt), verifying the strong representation ability of PTMs. This implies that more complex learning systems do not guarantee better performance, which can even introduce noise across incompatible modules.

\noindent\textbf{Discussions on comparison fairness}: From Table~\ref{tab:benchmark}, we observe prompt-based methods perform poorly except for DAP. However, we find a fatal problem in DAP, which could influence future comparison fairness. Specifically, DAP generates instance-specific prompts via Eq.~\ref{eq:dap}. However, the $\gamma_e, \beta_e$ in the equation rely on {\em voting of a same batch}. 
During inference, it clusters instances from the same task in the same batch and uses the same $\gamma_e, \beta_e$ generation for the same batch. In other words, it is equal to directly annotating the task identity and simplifying the difficulty. 
When we set the testing batch size to $1$, \ie, removing the batch information in DAP (denoted as DAP w/o BI), we observe a drastic degradation in the performance. DAP w/o BI even works inferior to typical prompt-based methods L2P, verifying that the core improvements come from the batch voting information. Since machine learning models should be tested independently, utilizing such context information obviously results in an unfair comparison.
In this paper, we would like to point out the unfairness and get the CL comparison back on track.

%% file: future.tex
\section{Future Directions}

\noindent\textbf{Continual learning with pre-trained large language models (LLMs)}:
In the current landscape dominated by PTMs, the capability for continual learning in LLMs like GPT~\cite{floridi2020gpt} is increasingly vital. These models need to adapt to ever-evolving information, such as changing global events. For instance, post the 2020 election, GPT required an update from `Who is the current president of the US?' $\rightarrow$ `Donald Trump' to `Joe Biden'. Typically, this would necessitate a comprehensive re-training with an updated dataset, as incremental fine-tuning might lead to overwriting other related knowledge. This process is resource-intensive, involving thousands of A100 GPUs over several months, incurring substantial electricity costs, and contributing to significant CO2 emissions.
CL presents a solution, enabling LLMs to be updated progressively with new concepts. This setting, often referred to as lifelong model editing in literature~\cite{zhang2024comprehensive}, shares methodologies with PTM-based CL. Thus, the development of CL strategies for LLMs represents a promising avenue for future research, potentially reducing resource consumption and enhancing the responsiveness of these models to current information.

\noindent\textbf{Beyond single modality recognition}: This survey primarily  focuses on the advancements of PTM-based CL in visual recognition, a key area in machine learning. However, the scope of recent progress in pre-training extends beyond single modality models to encompass multi-modal PTMs, such as CLIP~\cite{radford2021learning}. These multi-modal PTMs are capable of processing, responding to, and reasoning with various types of input.
While notable progress in visual recognition has been achieved, particularly in leveraging textual information to enhance and select appropriate prompts~\cite{khan2023introducing,villa2023pivot}, there is a burgeoning interest in expanding beyond visual recognition. For instance, PROOF~\cite{zhou2023learning} advances the continual learning capabilities of CLIP and other vision-language models for various multi-modal tasks. This is achieved by introducing a cross-modal fusion module, signifying a significant step forward in multi-modal continual learning. This shift towards integrating multiple modalities opens up exciting new pathways for future research and applications in the field.

\noindent\textbf{Learning with restricted computational resources}: The proficiency of large PTMs in various tasks is undeniable, yet the ongoing tuning of these models often incurs significant computational costs. In the context of PTMs, the deployment of models is not limited to cloud-based environments but extends to edge devices as well. 
A pertinent example is the training of LLMs for personal assistant smartphone applications, which demands local training and inference.
 This scenario necessitates continual learning algorithms that are computationally efficient. Reflecting this need, recent advancements in continual learning~\cite{prabhu2023computationally} have increasingly focused on scenarios with limited resources. This trend is likely to illuminate and address crucial challenges related to computational efficiency in future developments.

\noindent\textbf{New benchmarks beyond PTM knowledge}: The essence of CL is to equip a learning system with the ability to acquire knowledge it previously lacked. Nevertheless, given the extensive training datasets used for PTMs, such as ImageNet, these models seldom encounter unfamiliar information. Consequently, training PTMs on a subset of their pre-training dataset can be redundant. There's a growing need for new datasets that exhibit a significant {\em domain gap} compared to ImageNet to challenge these models effectively. In this survey, we follow~\cite{zhou2023revisiting} by utilizing ImageNet-R/A, ObjectNet, OmniBenchmark, and VTAB for evaluations. These datasets offer a diverse range of data with substantial domain gaps relative to ImageNet. However, as training techniques and datasets continue to evolve, identifying and leveraging new benchmarks that present novel challenges to PTMs --- data they have not previously encountered and must learn --- remains an intriguing and important direction.

\section{Conclusion} 

Real-world applications require the ability to continually update the model without forgetting. Recently, the introduction of pre-trained models has substantially changed the way we do continual learning. In this paper, we provide a comprehensive survey about continual learning with pre-trained models by categorizing them into three categories taxonomically. Moreover, we conduct extensive experiments on seven benchmark datasets for a holistic evaluation among methods from these categories. We summarize the results and raise a fair comparison protocol with batch-agnostic inference. Finally, we point out the future directions of PTM-based CL. We expect this survey to provide an up-to-date summary of recent work and inspire new insights into the continual learning field.

\section*{Acknowledgments}
This work is partially supported by National Science and Technology Major Project (2022ZD0114805),
NSFC (62376118, 62006112, 62250069, 61921006), Collaborative Innovation Center of Novel Software
Technology and Industrialization.